\documentclass{article} %
\usepackage{iclr2025_conference,times}

\usepackage{amsmath,amsfonts,bm}

\def\eqref#1{equation~\ref{#1}}

\def\1{\bm{1}}

\DeclareMathAlphabet{\mathsfit}{\encodingdefault}{\sfdefault}{m}{sl}
\SetMathAlphabet{\mathsfit}{bold}{\encodingdefault}{\sfdefault}{bx}{n}

\DeclareMathOperator*{\argmax}{arg\,max}
\DeclareMathOperator*{\argmin}{arg\,min}

\usepackage{hyperref}
\usepackage{amsmath,amssymb,amsthm,amstext,amsfonts}
\usepackage{mathrsfs}
\usepackage{algorithm, algorithmicx}
\usepackage{graphicx}
\usepackage{xspace}
\usepackage{nicefrac}
\usepackage{color}
\usepackage{enumerate}
\usepackage{bm}
\usepackage{wrapfig}
\usepackage{textcomp}
\usepackage{tabularx}
\usepackage{verbatim}
\usepackage{caption}
\usepackage{comment} 
\usepackage{epsfig} 
\usepackage{latexsym}
\usepackage{bbm}
\usepackage{nicefrac}
\usepackage{url}
\usepackage{subfigure}
\usepackage{enumitem}
\usepackage{booktabs}
\usepackage{commath}
\usepackage{mdframed}
\usepackage{pdfsync}

\def\balign#1\ealign{\begin{align}#1\end{align}}
\def\baligns#1\ealigns{\begin{align*}#1\end{align*}}
\def\balignat#1\ealign{\begin{alignat}#1\end{alignat}}
\def\balignats#1\ealigns{\begin{alignat*}#1\end{alignat*}}
\def\bitemize#1\eitemize{\begin{itemize}#1\end{itemize}}
\def\benumerate#1\eenumerate{\begin{enumerate}#1\end{enumerate}}

\newenvironment{talign*}
 {\csname align*\endcsname}
 {\endalign}
\newenvironment{talign}
 {\csname align\endcsname}
 {\endalign}

\def\balignst#1\ealignst{\begin{talign*}#1\end{talign*}}
\def\balignt#1\ealignt{\begin{talign}#1\end{talign}}
\let\originalleft\left
\let\originalright\right
\renewcommand{\left}{\mathopen{}\mathclose\bgroup\originalleft}
\renewcommand{\right}{\aftergroup\egroup\originalright}

\def\tinycitep*#1{{\tiny\citep*{#1}}}
\def\tinycitealt*#1{{\tiny\citealt*{#1}}}
\def\tinycite*#1{{\tiny\cite*{#1}}}
\def\smallcitep*#1{{\scriptsize\citep*{#1}}}
\def\smallcitealt*#1{{\scriptsize\citealt*{#1}}}
\def\smallcite*#1{{\scriptsize\cite*{#1}}}

\def\mrm#1{\mathrm{#1}}

\def\<{\left\langle} %
\def\>{\right\rangle}

\def\norm#1{\left\|{#1}\right\|} %
\newcommand{\binner}[2]{\left\langle{#1},{#2}\right\rangle} %

\providecommand{\argmax}{\mathop\mathrm{arg max}} %
\providecommand{\argmin}{\mathop\mathrm{arg min}}

\providecommand{\abs}{\mathop\mathrm{abs}}

\newcommand{\defn}[1]{\emph{#1}}

\newcommand{\loss}{\ell}

\ifdefined\nonewproofenvironments\else
\ifdefined\ispres\else
\newtheorem{theorem}{Theorem}

\newtheorem{definition}[theorem]{Definition}

\newenvironment{proof-sketch}{\noindent\textbf{Proof Sketch}
  \hspace*{1em}}{\qed\bigskip\\}
\newenvironment{proof-idea}{\noindent\textbf{Proof Idea}
  \hspace*{1em}}{\qed\bigskip\\}
\newenvironment{proof-of-lemma}[1][{}]{\noindent\textbf{Proof of Lemma {#1}}
  \hspace*{1em}}{\qed\\}
\newenvironment{proof-of-theorem}[1][{}]{\noindent\textbf{Proof of Theorem {#1}}
  \hspace*{1em}}{\qed\\}
\newenvironment{proof-attempt}{\noindent\textbf{Proof Attempt}
  \hspace*{1em}}{\qed\bigskip\\}

\newenvironment{proof-of-claim}[1][{}]{\noindent\textbf{Proof of Claim {#1}}
\hspace*{1em}}{\qed\\}

\fi
\fi
\makeatletter
\@addtoreset{equation}{section}
\makeatother

\DeclareMathOperator*{\av}{\mathbbm{E}}
\DeclareMathOperator*{\var}{\bf Var}
\DeclareMathOperator*{\std}{\bf std}

\renewcommand{\norm}[1]{\ensuremath{\left\lVert #1 \right\rVert}}

\newcommand{\marginlabel}[1]%
{\mbox{}\marginpar{\it{\raggedleft\hspace{0pt}#1}}}

\renewcommand\set[1]{\left\{#1\right\}} %

\newcommand{\ones}{\mathbbm{1}}

\global\long\def\Reals{\mathbb{R}}

\newcommand\calA{\mathcal{A}}
\newcommand\calF{\mathcal{F}}
\newcommand\calS{\mathcal{S}}

\newcommand\calD{\mathcal{D}}

\newcommand\calE{\mathcal{E}}

\newcommand{\sqparen}[1]{\unskip\left[{#1}\right]}

\newlength{\pgmtab}  %
 {
	\begin{enumerate}}{\end{enumerate}}

\newcommand\np{\mbox{\bf NP}\xspace}

\newcommand\EE[2]{\av_{#1}\sqparen{#2}}

\newcommand{\permuteop}[2]{#1[#2]}

\usepackage{multicol}
\usepackage{booktabs}
\usepackage{multirow}

\usepackage{url}
\usepackage{graphicx}
\usepackage{algorithm}
\usepackage{algpseudocode}
\usepackage[hide=true,marginparwidth=1.5in]{marginalia}
\usepackage[capitalize]{cleveref}
\definecolor{DarkBlue}{rgb}{0.075,0.05,0.6}
\usepackage{hyperref}
\hypersetup{
	linktocpage=true,
	colorlinks=true,				
	linkcolor=DarkBlue,				
	citecolor=DarkBlue,				
	urlcolor=DarkBlue,
}

\usepackage{fullpage}
\usepackage{wrapfig}

\newcommand{\traindata}[1]{\calD^{\mrm{train}}_{#1}}
\newcommand{\testdata}[1]{\calD^{\mrm{test}}_{#1}}
\newcommand{\nonlocal}{non-local\xspace}
\newcommand{\local}{local\xspace}
\newcommand{\Nonlocal}{Non-local\xspace}
\newcommand{\NonLocal}{Non-Local\xspace}
\newcommand{\merged}{\mrm{merged}}
\newcommand{\expertset}{\calE}
\newcommand{\init}{\mrm{init}}
\newcommand{\localized}{\mrm{localized}}

\newcommand{\MTL}{\mrm{MTL}}

\title{The \NonLocal Model Merging Problem:  \\ Permutation Symmetries and Variance Collapse}
\iclrfinalcopy
\author{Ekansh Sharma \\
University of Toronto; Vector Institute\\
\texttt{ekansh@cs.toronto.edu} \\
\And
Daniel M. Roy \\
University of Toronto; Vector Institute\\
\texttt{droy@utstat.toronto.edu} \\
\AND
Gintare Karolina Dziugaite \\
Google DeepMind \\
\texttt{gkdz@google.com}
}

\begin{document}

\maketitle

\begin{abstract}

Model merging aims to efficiently combine the weights of multiple expert models, each trained on a specific task, into a single multi-task model, with strong performance across all tasks.  When applied to all but the last layer of weights, existing methods---such as Task Arithmetic, TIES-merging, and TALL mask merging---work well to combine expert models obtained by fine-tuning a common foundation model, operating within a ``local'' neighborhood of the foundation model. This work explores the more challenging scenario of ``non-local'' merging, which we find arises when an expert model changes significantly during pretraining or where the expert models do not even share a common foundation model.

We observe that standard merging techniques often fail to generalize effectively in this non-local setting, even when accounting for permutation symmetries using standard techniques.  We identify that this failure is, in part, due to ``variance collapse'', a phenomenon identified also in the setting of linear mode connectivity by \citet{jordan2022repair}.
To address this, we propose a multi-task technique to re-scale and shift the output activations of the merged model for each task, aligning its output statistics with those of the corresponding task-specific expert models.  Our experiments demonstrate that this correction significantly improves the performance of various model merging approaches in non-local settings, providing a strong baseline for future research on this problem.

\end{abstract}

\section{Introduction}

The pursuit of state-of-the-art performance in machine learning currently involves fine-tuning a pretrained foundation model on a task-specific dataset.  This has led to a proliferation of foundation models and their task-specific adaptations, with significantly enhanced performance on downstream tasks \citep{wolf-etal-2020-transformers}.  However, deploying these models for real-world applications can be resource-intensive, as the process produces a dedicated model for each task.

Model merging offers a solution by combining expert models trained on different tasks into a single, multi-task model.  
Methods like Task Arithmetic \citep{ilharco2023editing} and TIES merging \citep{yadav2023tiesmerging} have shown promising results in recovering the performance of individual expert models.  
TALL-mask merging \citep{wang2024localizing}, a compression-based approach, has demonstrated the ability to maintain the bulk of task-specific performance, even when combining models from over 20 different tasks.

\begin{figure}[t]
    \centering
    \includegraphics[width=0.32\linewidth]{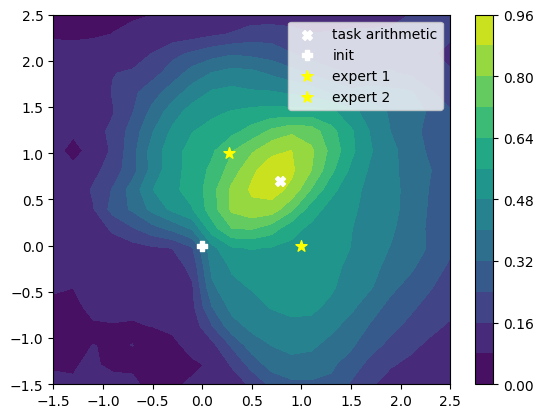}
    \includegraphics[width=0.32\linewidth]{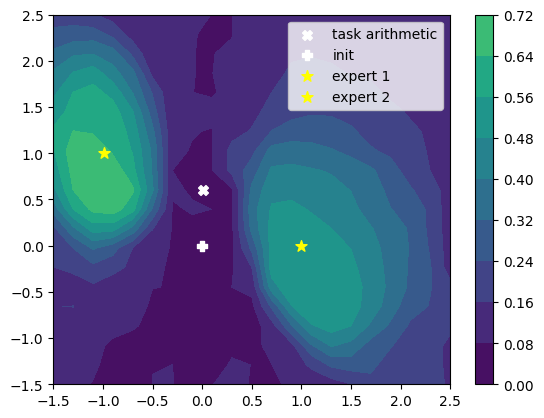}
    \includegraphics[width=0.32\linewidth]{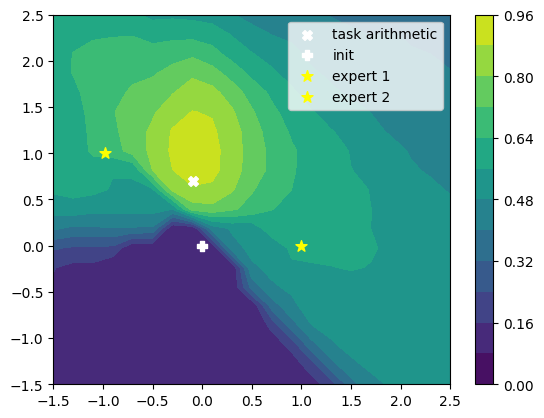}
    \caption{Local vs Non-Local Accuracy Landscape. For the VGG16 model architecture, we visualize the average normalized accuracy landscape in the span of the task vectors (treating the initialization as the origin) for two classification tasks: RESISC45 and Colorectal Histology. 
    \textbf{(Left)} Local merging: Expert models share the same fine-tuning initialization; 
    \textbf{(center)} \Nonlocal merging: Expert models are derived from different pretrained models;
    \textbf{(Right)} \Nonlocal merging modulo permutation and TACT: Expert models are derived from different pretrained models and merged modulo permutation, followed by
    task-specific activation repair. 
    }
    \label{fig:avg-loss-local-v-non-local}
\end{figure}

Existing model merging methods work effectively when merging expert models fine-tuned from a common foundation model. In this case, the expert models typically lie in the ``local'' neighborhood of the common foundation model (illustrated in \cref{fig:avg-loss-local-v-non-local}, left panel).  This is not always the case, however. 
When expert models are generated by fine-tuning independent foundation models we have observed poor performance by existing model merging methods. We call such merging scenarios ``non-local''.

In this work, we ask: Are there reliable approaches to non-local model merging? Our goal is to develop a strong baseline that we hope will spur further research. 

In the single-task setting, linear (mode) connectivity \citep{frankle2020linear,nagarajan2019uniform} is the phenomenon in deep learning where, under certain conditions, independent training runs produce weights whose convex combinations (in other words, naive merges) perform essentially as well. 
Recent research \citep{entezari2021role,ainsworth2022git,singh2020model,sharma2024simultaneous,jordan2022repair}, inspired by a conjecture, preliminary empirical work, and theoretical evidence for wide shallow networks by \citet{entezari2021role},
has shown that, in a much wider array of situations, pairs of trained networks are linearly connected \emph{modulo permutation}, i.e., provided one applies a carefully chosen permutation to one of the networks before evaluating linear connectivity.
Recent work provides empirical evidence that a single permutation connects entire training trajectories \citep{sharma2024simultaneous}.

These recent findings on linear connectivity modulo permutation raise the possibility that we may be able to address the non-local model merging problem simply by accounting for permutations. Our experiments do not bear this out: after merging the permuted models, the merged model produces significantly different representations, which is evident even in the scale and variance of the activations.
A similar variance problem was identified by \citet{jordan2022repair}, while studying linear connectivity in networks with batch-norm layers.
They proposed to adjust the batch-norm layers, after merging, to account for the variance collapse that occurs from merging.
We extend this idea beyond the single-task setting to a multitask setting. 
Our results demonstrate that non-local merging can be successful when accounting for permutations and appropriately rescaling the merged network so that the statistics of its activations match those of the original expert model.
In fact, we show that the same fix also improves local multitask merging.

\begin{figure}[t]
    \centering
    \includegraphics[width=0.32\linewidth]{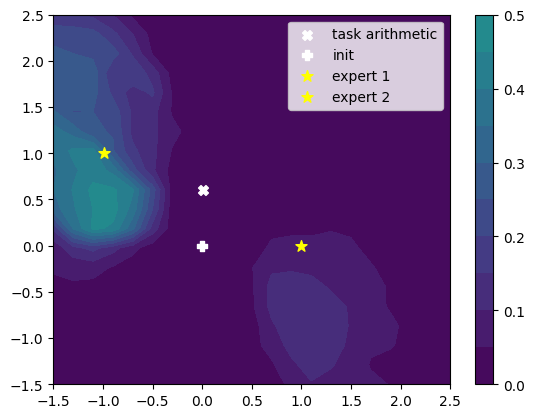}
    \includegraphics[width=0.32\linewidth]{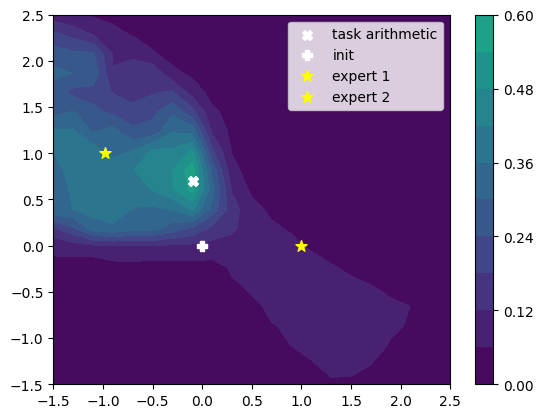}
    \includegraphics[width=0.32\linewidth]{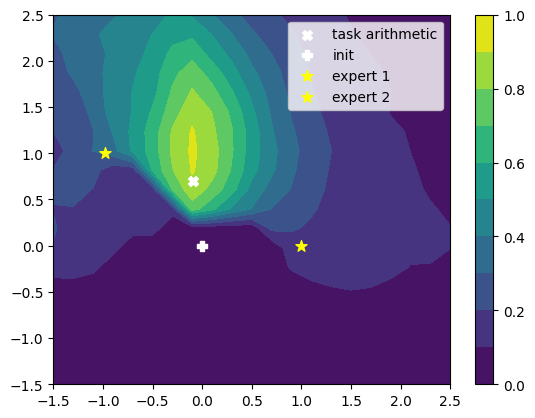}
    \caption{Local vs Non-local Worst Case Landscape. For the same architecture and tasks as in \cref{fig:avg-loss-local-v-non-local}, we visualize the worst-case accuracy landscape, i.e., for every setting of weights in the hyperplane, the color conveys the performance of task with the \
    \emph{lowest} normalized accuracy. \textbf{(Left)} No permutation: Using the average of the two foundation models as initialization, $\theta_\init$, we look at the worst case accuracy in the span of the two task vectors $\tau_t = \theta_t-\theta_\init$. Worst case accuracy is low throughout. \textbf{(Center)} Modulo permutation: Using the average of  two foundation models modulo permutations as initialization, we plot the worst-case accuracy along the permuted task vectors. We notice a moderate improvement in worst case performance;   \textbf{(Right)} Modulo permutation + TACT: Like the center plot, but the TACT correction is applied before evaluating accuracy, resulting in a significant improvement to worst case accuracy. }
    \label{fig:min-acc-non-local-mod-p}
\end{figure}

\paragraph{Contributions.}
\begin{enumerate}[leftmargin=1.5em]
    \item We introduce the problem of multi-task \nonlocal model merging. This is the general problem of merging expert models, with the same architecture, but trained on different tasks, without requiring them to be derived from the same foundation model.
    \item We extend existing model merging algorithms to the non-local setting. This is achieved by incorporating a permutation alignment step to align the networks before merging.
    \item We propose a new algorithm, Task-specific Activation Correction (TACT), to improve merging performance. TACT employs layer-wise rescaling and shifting to align the activations of the merged model with the expert models. It significantly improves multi-task model merging performance with minimal memory overhead (e.g., only 0.4\% additional parameters for ViTB/16).
    \item We demonstrate the effectiveness of our approach in both \nonlocal and local merging scenarios. TACT consistently improves performance when merging models from different foundation models (non-local) and even enhances merging with weak foundation models in the local setting.
\end{enumerate}

\section{Background}

\subsection{Linear connectivity modulo permutation}
\label{sec:lcmp}
Linear mode connectivity (LMC) is a property of loss landscapes that has been extensively studied in machine learning.  
The concept of LMC was first introduced and named by  \citet{frankle2020linear}, inspired by a related observation by \citet{nagarajan2019uniform}.  \citet{frankle2020linear} showed the existence of linear paths between independently trained neural networks such that all weights along the path parameterize a network that achieves around the same accuracy on a given task as the average of the end-points. More formally, if $\theta_1$, $\theta_2$ are parameters of a pair of networks, the error barrier is defined as
\[
\sup_{\alpha \in (0,1) } \loss( \alpha \theta_1 + (1-\alpha) \theta_2) - \alpha \loss(\theta_1) - (1-\alpha)  \loss(\theta_2),
\]
where $\loss$ is the loss function of interest evaluated on the training or test data (depending on what landscape we want to measure the barrier in). 
We say $\theta_1$ and $\theta_2$ are linearly connected if the error barrier is ``close to zero''. (The Monte Carlo noise floor is one natural threshold.)

Linear mode connectivity is a useful property because it allows for the interpolation between different models without sacrificing performance.  This can be used for various applications, such as model merging, knowledge transfer, and model compression. In the context of model merging, linear mode connectivity enables the combination of two or more trained models into a single model that performs well on all tasks. 

\citet{frankle2020linear} observed that, when Stochastic Gradient Descent (SGD) or its variants are used to train neural networks, there was a point, early in training, when networks always converged to the same linearly connected mode, regardless of the stochasticity in the minibatches, etc. For most large, trained networks, this point in training was not the start of training: separate runs resulted in networks separated by a high loss barrier along their convex combination. 

\paragraph{Permutation symmetries.} Neural networks exhibit \defn{permutation symmetries}, meaning that intermediate neurons of a network can be permuted without functionally altering the network. 
Building upon \citet{frankle2020linear}'s observation, \citet{entezari2021role} noticed that a network and its random permutation, while being functionally equivalent, were essentially never linearly connected, leading them to conjecture that permutations were the only source of error barriers in trained networks.
They conjectured that despite the barriers, all SGD-trained networks converge to a single basin modulo permutations.  This conjecture suggests that by considering permutations of neurons, the seemingly disconnected solutions can be brought into the same basin and potentially connected with a low-loss linear path. 
Subsequent research has provided empirical support for this conjecture, demonstrating that these error barriers can indeed be reduced between two independently trained networks by employing permutation alignment algorithms \citep{wang2020federated,singh2020model,ainsworth2022git}.
These algorithms were designed to efficiently look for a permutation that decreases the distance between activations or weights between two networks. While the search is imperfect, the algorithms do succeed in aligning networks 
under certain conditions, such as using extra wide networks or networks with layer normalization \citep{sharma2024simultaneous,jordan2022repair}. 

\paragraph{Variance collapse and activation repair.}

Linear interpolation between networks, especially in the context of permutation-based approaches, can lead to a phenomenon known as variance collapse.  This involves a significant reduction in the variance of activations in the interpolated model compared to the individual expert models.  Such a collapse is especially problematic in deep networks. In general, generalization suffers as the degree of collapse increases.

To counter this, \citet{jordan2022repair} proposed a technique called activation repair, aimed at restoring the variance of activations in the merged model. 
Their method involves rescaling and shifting the activations to match the average statistics of the individual expert models. Let $\calA^{(k)}_{\theta_1}(X)$,$\calA^{(k)}_{\theta_2}(X)$  denote the channel activations of $k$\textsuperscript{th} module of the end points $\theta_1, \theta_2$. 
They propose to correct the channel-wise activations of the merged model, $\theta_\alpha$ to satisfy:
\begin{align}
    \av_{X\sim \calD}[\calA^{(k)}_{\theta_\alpha}(X)]& =  (1-\alpha) \cdot \av_{X\sim \calD}[\calA^{(k)} _{\theta_1}(X)] + \alpha \cdot \av_{X\sim \calD}[\calA^{(k)} _{\theta_2}(X)] , \\
    \std_{X\sim \calD}[\calA^{(k)}_{\theta_\merged}(X)]& =  (1-\alpha) \cdot \std_{X\sim \calD^t}[\calA^{(k)}_{\theta_1}(X)]  +  \alpha \cdot \std_{X\sim \calD^t}[\calA^{(k)}_{\theta_2}(X)].
\end{align}
This can be achieved by incorporating Batchnorm layers after each module experiencing variance collapse, effectively renormalizing the activations and restoring the lost variance.  In doing so, they improve the performance of the linearly interpolated model.

\subsection{Model merging}
\label{sec:merging}

Model merging is a machine learning technique that aims to combine multiple models, usually trained on different tasks, into a single model capable of performing well on all those tasks. This approach offers several potential benefits, such as reduced memory footprint, simplified deployment, and improved performance, especially in resource-constrained environments. 

\begin{definition}[The \local merging problem]
Let $\theta_{\mathrm{init}}$ denote the weights of a pre-trained foundation model.
For each $t \in [T]$,
let $(\traindata{t}, \testdata{t})$ be i.i.d.\ task-specific training and test datasets, and let $\theta_t$ be an expert model obtained by fine-tuning $\theta_{\mathrm{init}}$ on $\traindata{t}$.
The local merging problem is to find a function  $\mathrm{LocalMerging}$ that combines
$\theta_1, \dots, \theta_T$ into a single model $\theta_{\mrm{merged}} = \mathrm{LocalMerging}(\theta_{\mathrm{init}},\{\theta_t\}_{t})$, such that $\theta_{\mrm{merged}}$ generalizes well across all $\testdata{t}$. 
\end{definition}

Standard model merging functions $\mathrm{LocalMerging}$ work by first finding a so-called \defn{task vectors}, defined as $\tau_t = \theta_t - \theta_{\init}$, then potentially processing it further via, e.g., masking, rescaling, etc., and eventually averaging to obtain $\tau_{\MTL}$. 
The merged model is then obtained by 
\[
\theta_\merged = \theta_\init + \tau_\MTL.
\]

We review standard techniques below.

\paragraph{Average merging.}
Average-based merging is a straightforward approach to model merging, where the weights of the individual expert models $\theta_t$ are simply averaged, $
\theta_\merged := \frac1T\sum_t\theta_t$.
This technique has been shown to be surprisingly effective in some cases and can be linked to the concept of linear mode connectivity. 
If the expert models are located in the same (multi-task) loss basin and are linearly connected, averaging their weights can be seen as finding a point along the linear path connecting them.
There are  several variations and extensions of the basic approach. Some studies have explored weighted averaging, where different weights are assigned to different expert models based on their performance or importance. 
Other approaches have investigated layer-wise averaging, where the weights of each layer are averaged separately, allowing for more flexibility in combining models with different architectures.

\paragraph{Task arithmetic.}

Task arithmetic is a more recent model merging approach that proposed working with the task vectors $\tau_t$'s, instead of the $\theta_t$'s \citep{ilharco2023editing}. 
These task vectors are then averaged and added to or subtracted from the foundation model to produce a merged model with desirable properties, i.e., $
 \tau_\MTL  := \lambda \cdot \sum_t\tau_t$. 
This method has been shown to be more effective than simple averaging, particularly when merging models trained on diverse tasks.

\paragraph{Ties merging.}
TIES-merging, introduced by \citet{yadav2024ties}, is a model merging approach that specifically addresses the issue of interference between tasks when merging models.  
It recognizes that simply averaging or arithmetically combining model weights can lead to negative interference, where the performance on individual tasks deteriorates after merging.  
To mitigate this, TIES-merging employs a technique to identify and resolve interference by removing smallest magnitude weights on the task vectors, and only averaging the weights that agree in their sign (where the chosen sign is determined by averaging the task vectors).

\paragraph{TALL-mask merging.}
TALL-mask merging is a recent advancement in model merging that focuses on compressing the merged model based on a task-specific pruning mask \citep{wang2024localizing}.  It identifies and retains only the essential task-specific information from each expert model based on expert weight magnitudes, discarding small magnitude weights which have been shown in the literature to be largely redundant or unnecessary \citep{frankle2018lottery,zhu2017prune}: 
\[
\tau_\MTL := m_t \circ \tau'_\MTL, m_t = \ones\{\abs{\tau_t} \geq \abs {\tau_\MTL - \tau_t}\cdot \lambda_t\}, 
\]
where $\tau'_\MTL$ is the merged task vector obtained by any other merging approach

\section{The \nonlocal model merging problem}

We begin by defining the \nonlocal merging problem. 

\begin{definition}[The \nonlocal merging problem]
Consider a set  $\calF$ of pretrained foundation models obtained by independently training on a large dataset. 
Given task-specific training and test datasets $(\traindata{t}, \testdata{t})$ for $t\in [T]$, we obtain an expert model $\theta_t$ for each task $t$ by selecting a pretrained model $\theta_0 \in \calF$ and fine-tuning it on the corresponding $\traindata{t}$.
The goal of the \nonlocal merging problem is to merge the weights of the expert models $\theta_1, \dots, \theta_T$ into a single model $\theta_{\mrm{merged}}$ that generalizes well across all $\testdata{t}$.
\end{definition}

\Cref{fig:avg-loss-local-v-non-local} (left) illustrates that when expert models are finetuned from the same pretrained model, they remain in the same local linearly-connected basin. 
In contrast, when finetuned from different pretrained models, they land in different loss basins (\cref{fig:avg-loss-local-v-non-local} middle).
Based on the conjecture by \citet{entezari2021role} saying that SGD-trained networks are linearly connected modulo permutation, one may hope to solve the \nonlocal merging problem via permutation alignment, which, as discussed in \cref{sec:lcmp}, may enable us to align models from different basins.

What we show in this section is that accounting for permutation symmetries is \emph{not} enough.
Even with standard known tricks \citep{ainsworth2022git} to improve model alignment after permutation, the performance gap persists.

\subsection{Localizing model merging with permutations}

In this section we present an algorithm for combining model merging techniques discussed in \cref{sec:merging} with permutation alignment algorithms, and evaluate this combination in a \nonlocal setting.

Existing algorithms for finding permutations have limitations, and only work in specific settings.
Further, as pointed out by \citet{sharma2024simultaneous}, prior work only provides evidence for the so-called \defn{weak linear connectivity}, meaning that only two networks at a time can be aligned. 
Therefore, to evaluate whether current permutation algorithms may solve the \nonlocal merging problem, 
we constrain the problem by focusing on $\calF$ such that $\calF$ are obtained by independently training on the same large pretraining dataset, and $|\calF| = 2$.
This simplification allows us to explore the challenges of non-local merging while working within the capabilities of current permutation algorithms.

\begin{algorithm}
\caption{\Nonlocal model merging with permutations}
\label{alg:nlocal-merging}
\hspace*{\algorithmicindent}\textbf{inputs}: Foundation models $\calF= \set{\theta_0, \theta_1}$;   Expert models $\expertset = \set{\theta_1, \dots, \theta_T}$ 
\begin{algorithmic}[1]
\State $P^\star \gets \arg\inf_{P \in \mathcal{S}} d(\theta_0, \permuteop{P}{\theta_1})$ \Comment{Localize foundation models using permutations. } 
\State $\bar\theta_{\init} \gets \frac{\theta_0 + \permuteop{P^\star}{\theta_1}}{2}$  \Comment{Set the initialization by averaging the permuted foundation models}
\State $\expertset_{\mrm{localized}} \gets \set{\theta : \theta \in \expertset\text{ finetuned from }\theta_0} \cup \set{\permuteop{P^\star}{\theta} : \theta \in \expertset \text{ finetuned from }\theta_1}$ 
\State  $\theta_\merged \gets \mrm{LocalMerging}(\bar\theta_{\init}, \expertset_{\localized})$ 
\end{algorithmic}
\hspace*{\algorithmicindent}\Return $\theta_\mrm{merged}$
\end{algorithm}

In the \nonlocal merging problem $\theta_{\init}$ is not shared among the experts, and $\theta_t$'s are thus not aligned. 
In \cref{alg:nlocal-merging} we first align a pair of foundation models by computing a permutation that brings the two models to the same linearly connected loss basin. Averaging these localized initializations returns $\bar\theta_\init$.
We then apply this same permutation to finetuned expert models, yielding a set of localized expert weights $\expertset_\localized$.
Post-localization, we apply standard $\mathrm{LocalMerging}$ functions on the localized $\bar\theta_\init$ and $\expertset_\localized$.

\paragraph{Permutations alone do not close the performance gap.}

We observe that the average normalized accuracy in the \nonlocal merging setting (\cref{tab:non-local-merging}) is significantly lower than the average normalized accuracy in the local setting (\cref{tab:local-merging}) for the same model architecture VGG16 on same tasks. 

Loss landscape analysis reveals that the merged model post-alignment falls outside a minimum.
\Cref{fig:min-acc-non-local-mod-p} presents a visualization of the test accuracy landscape for the worst-performing task (after applying \cref{alg:nlocal-merging}).
The plot shows that our localization scheme leads to moderate improvement in the worst case loss but is insufficient to establish a successful merging procedure.

\paragraph{Merging modulo permutations negatively effects activations.}

\begin{figure}
    \centering
    \includegraphics[width=0.32\linewidth]{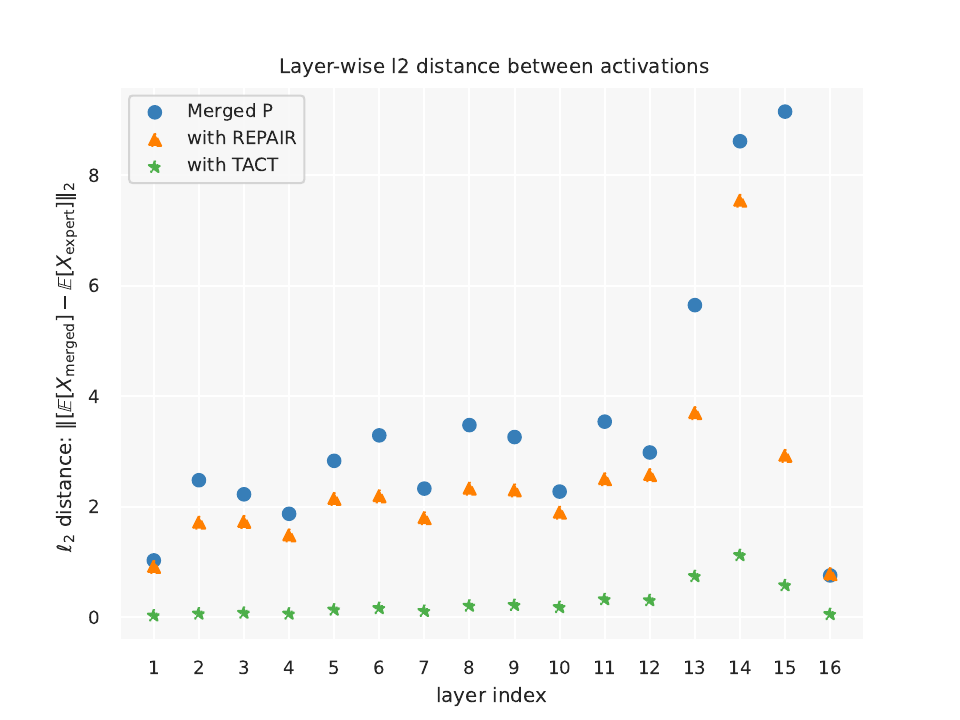}
    \includegraphics[width=0.32\linewidth]{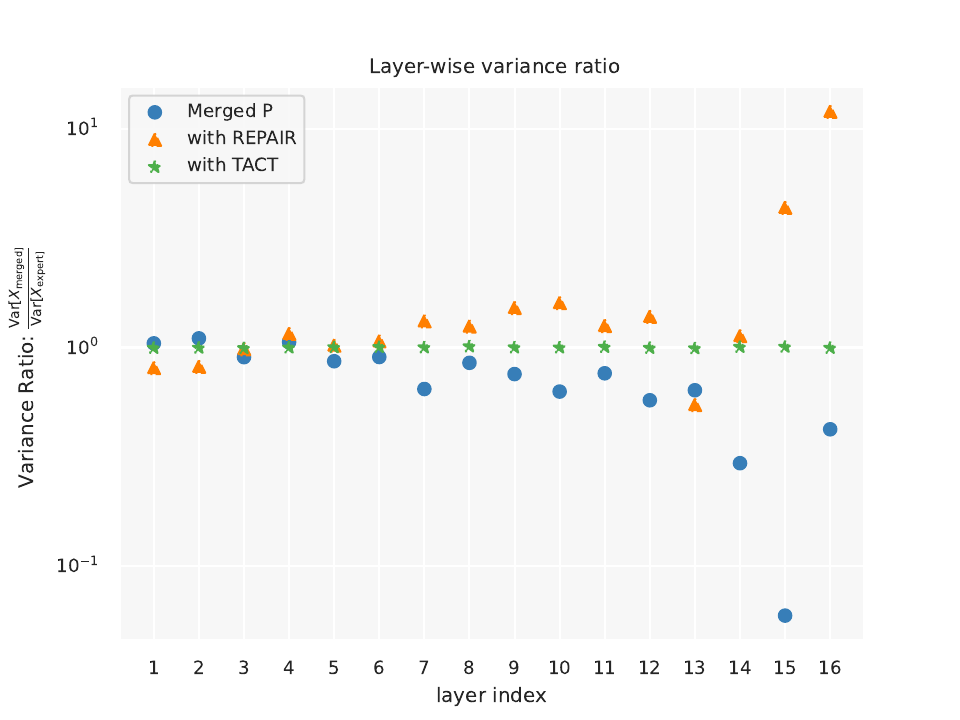}
    \caption{\textbf{Internal statistics of a merged model modulo permutations.} We plot the layer-wise activation statistics after merging 4 tasks using Task Arithmetic. We use of a batch of Colorectal Histology dataset for computing the statistics.  \textbf{(Left)} Per-layer $\ell_2$ distance between the activation vector of a merged model on task $t$, and activation vector of expert model for task $t$. \textbf{(Right)} The ratio of activation vector variance between the merged model on task $t$ to the expert model on task $t$, computed for each layer. We find that the internal statistics of the merged model on task $t$ do not match the internal statistics the expert model for task $t$.}
    \label{fig:internal-stats}
\end{figure}

To assess the impact of permutation alignment on model merging, we analyze the per-layer activation statistics of the merged model compared to the individual expert models. \Cref{fig:internal-stats} presents two key observations. 

First, as the layer index increases, the distance between the activations of the merged model and the expert model grows significantly (\Cref{fig:internal-stats} left).  
This indicates that the deeper layers of the merged model deviate substantially from the corresponding expert model, potentially hindering its ability to capture task-specific features. 
 
Second, the variance ratio (the ratio of per-layer activation variance between the merged model and the expert model) decreases with increasing layer index (\cref{fig:internal-stats} right).  This suggests that the merged model suffers from a reduction in activation variance compared to the expert models, which indicates some degree of  ``collapse'' in the representations, and could limit the merge model's expressiveness and generalization.  

\citet{jordan2022repair} proposed the REPAIR to fix the issue of variance collapse when interpolating between two networks trained on the same data distribution. One could ostensibly construct a mixture data distribution that encompasses all the tasks and use that to apply REPAIR. 
In \Cref{fig:tact-repair-lp-ablation} (center), we use this approach and apply REPAIR to the merged model. We find that applying REPAIR improves the performance of model merging when merging few models, however, it exhibits a significant slump in the performance when we try and merge more tasks. 
In \Cref{fig:internal-stats} (right) we see that after applying REPAIR, variance of the merged model explodes.

What these results highlight is that merging modulo-permutation returns penultimate layer representations that are distinct from the expert ones.
This may make the existing classification layer unsuitable for the new representations, potentially leading to the drop in accuracy. 
We directly evaluate this hypothesis in \cref{fig:tact-repair-lp-ablation} (right).
We show that retraining the classifier on the merged-modulo-permutations model slightly improves the performance, but still underperforms relative to the corresponding local merging baseline.

These findings highlight the limitations of relying solely on permutation alignment for the \nonlocal merging and motivate our approach in the \cref{sec:tact}, which directly addresses the variance issue and aims to align the representations more effectively.

\section{Task specific activation repair}
\label{sec:tact}
\begin{figure}
    \centering
    \includegraphics[width=0.32\linewidth]{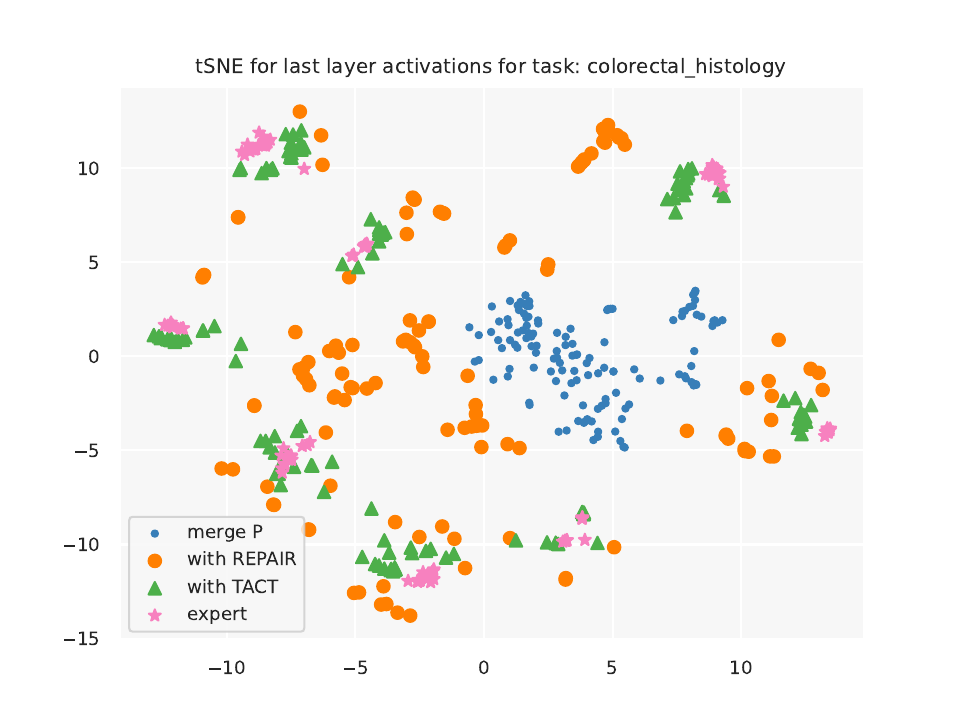}
    \includegraphics[width=0.32\linewidth]{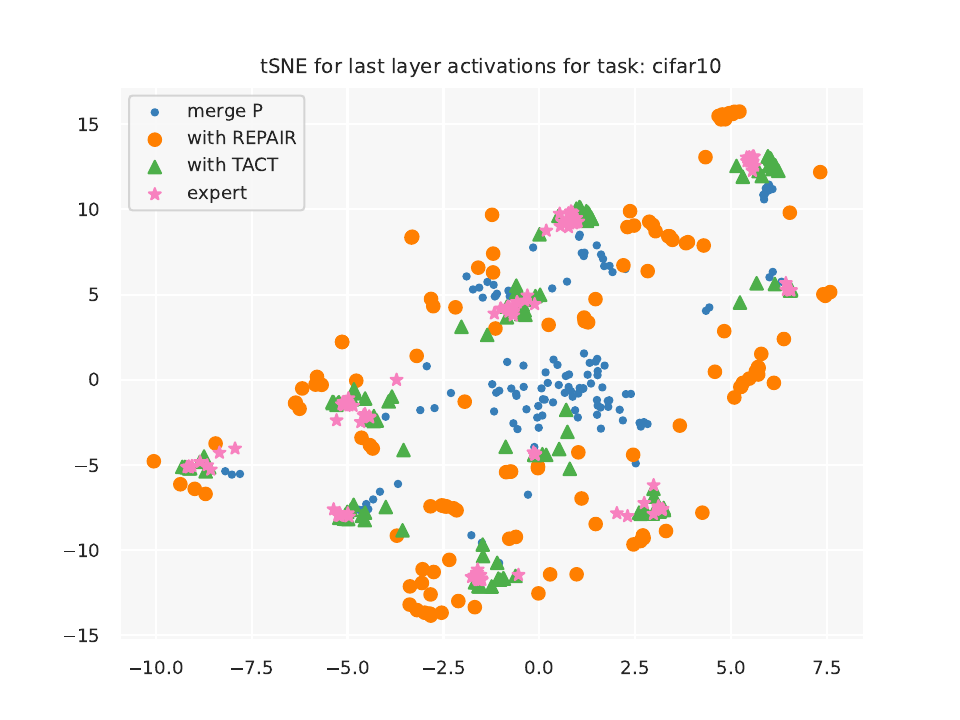}
    \caption{\textbf{t-SNE projection of the last layer embeddings.} For a batch of data, we visualize the t-SNE projections of the embeddings encoded by different merging methods models.  \textbf{(Left)} Colorectal Histology, \textbf{(Right)} CIFAR10. We observe that embeddings after TACT correction are closer to the embedding of the expert model.}
    \label{fig:tact-repair}
\end{figure}

To address the limitations of permutation alignment and the observed discrepancies in activation statistics, we propose a novel approach called Task-specific Activation Correction (TACT). TACT aims to improve non-local model merging by explicitly correcting the internal statistics of the merged model to better match those of the expert models.

Our method is inspired by the activation repair technique introduced by \citet{jordan2022repair} to mitigate variance collapse. When making inference on task $t$, we propose to correct the activations of the merged model to match those of the expert model for task $t$.  
In particular, let $\calA^{(k)}_{\theta_t}(X)$ denote the channel activations of $k$\textsuperscript{th} module of the $t$\textsuperscript{th} expert model on input data $X$. 
We propose to correct the channel-wise activations of the merged model to satisfy:
\begin{align}
    \av_{X\sim \calD^t}[\calA^{(k)}_{\theta_\merged}(X)]& =  \av_{X\sim \calD^t}[\calA^{(k)}_{\theta_t}(X)] ,\\
    \var_{X\sim \calD^t}[\calA^{(k)}_{\theta_\merged}(X)]& =  \var_{X\sim \calD^t}[\calA^{(k)}_{\theta_t}(X)] .
\end{align}
Similar to \cite{jordan2022repair}, we make this correction by incorporating BatchNorm layers into the model architecture, specifically after each module where activation statistics need correction. For each task $t$, we compute the activation means and variances for these modules and use these statistics to derive the BatchNorm parameters.  These parameters enable us to rescale and shift the activations of the merged model, effectively aligning them with the corresponding expert model. We detail our algorithm in \Cref{app:tact}.

In \cref{fig:tact-repair}, we visualize the internal layerwise activations with and without TACT. We observe that TACT allows the merged model to mimic the behaviour of the expert model and thus leading to improved model performance.   

\section{Empirical Evaluation}

We evaluate the performance of different merging approaches in the local and \nonlocal setting. 

For our evaluation we restrict ourselves to the vision model architectures including VGG16~\citep{Simonyan15} and ViTB16~\citep{dosovitskiy2021an}, pretrained with Imagenet-1k~\citep{ILSVRC15} and  Imagenet-21k~\citep{ridnik2021imagenet21k}. 
We finetune a pretrained model on a diverse set of classification tasks with data from various domains:  Natural images (CIFAR-10, CIFAR-100)~\cite{krizhevsky2009learning}, Scene understanding (SUN397~\citep{xiao2010sun}), plant images (Cassava~\citep{cassava-disease}, Flowers102), satellite imaging (EuroSAT~\citep{helber_eurosat_2019}, RESISC45~\citep{cheng17resisc}), Medical imaging (Colorectal Histology~\citep{kather_2016_53169}), pet images (Stanford Dogs~\citep{KhoslaYaoJayadevaprakashFeiFei_FGVC2011}, Oxford IIIT Pet~\citep{parkhi12a}), food images (Food101~\citep{bossard14}), and house images (SVHN~\citep{netzer11}).

To create each task-specific expert, we first train the classifier layer using the encoder of one of the pretrained models. 
Then, we finetune all layers of the expert model, initializing the encoder layers with the foundation model, and the classifier layer with the pretrained classifier.
For evaluation of the merged model for some task $t$, we reuse the classifier layer expert model for task $t$. The training hyperparameters for all models as well as the expert model accuracies are detailed in \cref{app:exp-details}.

\subsection{\Nonlocal model merging}
\paragraph{Setup.} 
We evaluate \nonlocal merging on VGG16 model architecture modulo permutation. We first independently train two models on Imagenet-1k dataset.
We use \defn{weight matching algorithm}~\citep{ainsworth2022git} to compute a permutation that aims to minimize the $\ell_2$ distance between the two pretrained networks. For posterity we give the description of the algorithm in \Cref{app:weightmatching}.

In our experiments, we merge an even number of experts, ensuring an equal number of experts are derived from each pretrained model to prevent bias towards a single mode.

\paragraph{Results.}

\begin{table}[]
    \centering
    \caption{Quantitative results for \nonlocal model merging for VGG16 when merging different number of tasks. We report average normalized accuracy and normalized accuracy corresponding to the worst task for each merging method. }
    \label{tab:non-local-merging}
\resizebox{0.9\textwidth}{!}{
\begin{tabular}{@{}cccccccccccccccccccccccccc@{}}
\toprule
\multicolumn{2}{c}{} &  & \multicolumn{6}{c}{VGG16} & &  \multicolumn{1}{c}{} \\ %
\cmidrule(l){3-10} %
\multicolumn{1}{c}{} & Merge Method  & \multicolumn{2}{c}{4 tasks} &  & \multicolumn{2}{c}{8 tasks} &  & \multicolumn{2}{c}{12 tasks} \\%& &  \multicolumn{1}{c}{4 tasks} &  & \multicolumn{1}{c}{8 tasks} &  & \multicolumn{1}{c}{12 tasks} & \\
\cmidrule(lr){3-4} \cmidrule(lr){6-7} \cmidrule(l){9-10} 
\multicolumn{1}{c}{} &  & Avg Acc. \% & Min Acc. (\%) \ &  & Avg Acc. \% & Min Acc. (\%) & & Avg Acc. \% & Min Acc. (\%) \\
\midrule
&Weight averaging &  43.35\% & 19.22\% &  & 35.26\% &  14.74\% &  & 29.75\% & 6.84\% \\%& 6 &  7 & 8 & 9 & 10 & 11 \\

&  Task arithmetic &  45.76\% & 18.83\% &  & 36.15\% & 16.91\% &   & 30.18\% & 6.63\% \\ %
&  TIES &  9.65\% & 1.40\%& & 12.63\% & 1.36\% &  & 8.05\% & 0.41\%\\%& 6 &  7 & 8 & 9 & 10 & 11 \\

 &  {TALL Mask + TA } &  58.37\% & 36.65\% &  & 55.51\% & 29.98\%&  & 42.39\% & 12.96\% \\ %

\multirow{-6}{*}{\rotatebox[origin=c]{90}{modulo P}} &
  {TALL Mask + TIES} & 68.74\% & 49/05\% &  & 57.59\% & 36.90\% &  & 43.41\% & 6.62\% \\%& 6 &  7 & 8 & 9 & 10 & 11 \\ 
 \midrule
 &Weight averaging &  55.70\% & 37.77\% & & 30.44\% & 12.51\% &  & 6.83\% & 0.39\%  \\ %

&  Task arithmetic &  44.72\% & 32.98\%  & & 30.44\% & 68.67\% &   & 5.10\% & 0.41\% \\%& 6 &  7 & 8 & 9 & 10 & 11 \\
&  TIES &  25.81\% & 12.29\% & & 7.09\% & 1.21\% &   & 5.10\% & 0.41\% \\%& 6 &  7 & 8 & 9 & 10 & 11 \\

 &  {TALL Mask + TA } &  59.81\% &  50.07\% & & 10.96\% & 2.19\% &  & 5.10\% & 0.42\% \\%&  &  7 & 8 & 9 & 10 & 11 \\

\multirow{-6}{*}{\rotatebox[origin=c]{90}{REPAIR}} &
  {TALL Mask + TIES} &  92.81\% & 87.03\% & & 89.25\% & 82.21\% &  & 80.27\% & 59.85\%  \\%& 6 &  7 & 8 & 9 & 10 & 11 \\ 

 \midrule
 &Weight averaging &  76.80\% & 60.17\% & & 66.68\% & 35.25\% &  & 55.25\% &18.65\%  \\ %

&  Task arithmetic &  85.89\% & 77.93\%  & & 81.26\% & 68.67\% &   & 70.18\% & 38.53\% \\%& 6 &  7 & 8 & 9 & 10 & 11 \\
&  TIES &  42.37\% & 22.01\% & & 33.76\% & 3.34\% &   & 27.49\% & 0.81\% \\%& 6 &  7 & 8 & 9 & 10 & 11 \\

 &  {TALL Mask + TA } &  \textbf{92.93\%} &  \textbf{88.08\%} & & \textbf{92.84\%} & \textbf{85.72\%} &  & \textbf{88.23\%} & \textbf{74.35\%} \\%&  &  7 & 8 & 9 & 10 & 11 \\

\multirow{-6}{*}{\rotatebox[origin=c]{90}{TACT}} &
  {TALL Mask + TIES} &  92.81\% & 87.03\% & & 89.25\% & 82.21\% &  & 80.27\% & 59.85\%  \\%& 6 &  7 & 8 & 9 & 10 & 11 \\ 
 \bottomrule
\end{tabular}
}

\end{table}

\begin{figure}[t]
    \includegraphics[width=0.32\linewidth]{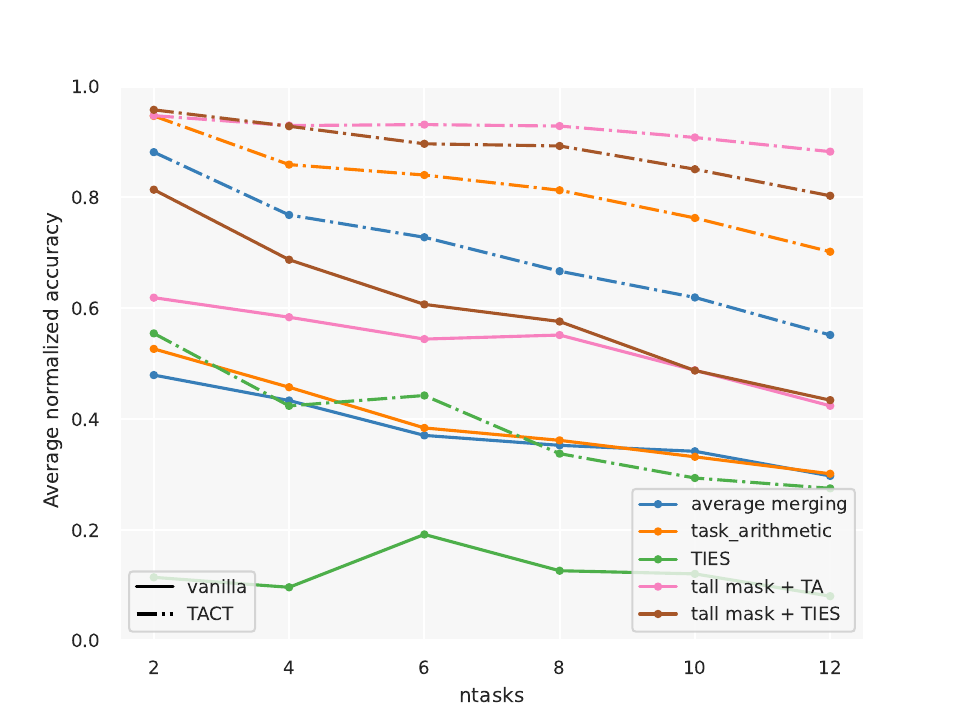}
    \centering \includegraphics[width=0.32\linewidth]{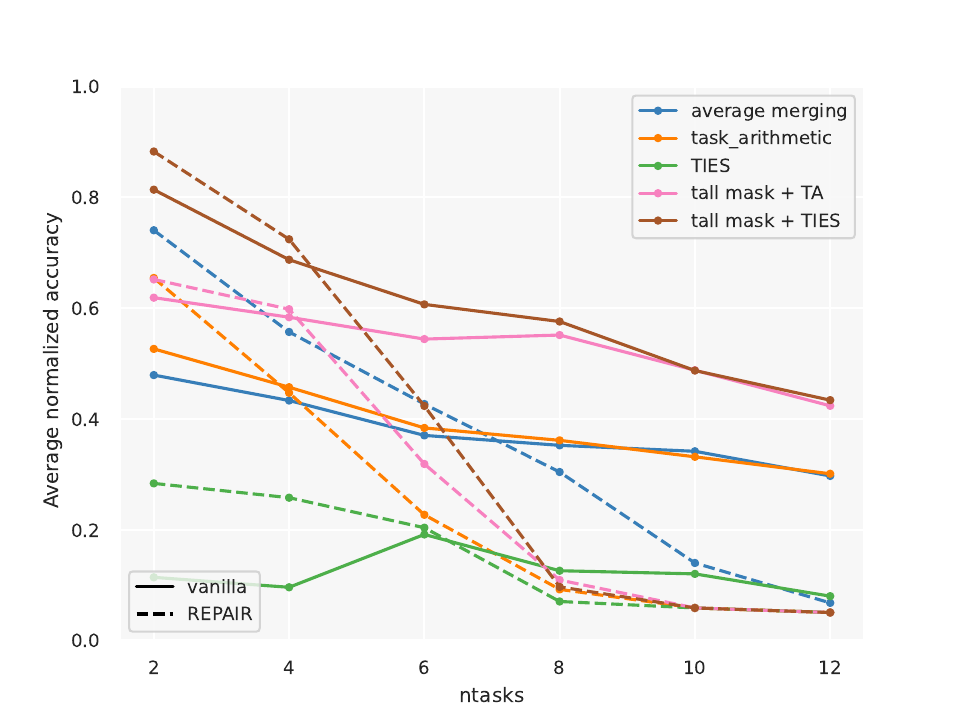}
    \includegraphics[width=0.32\linewidth]{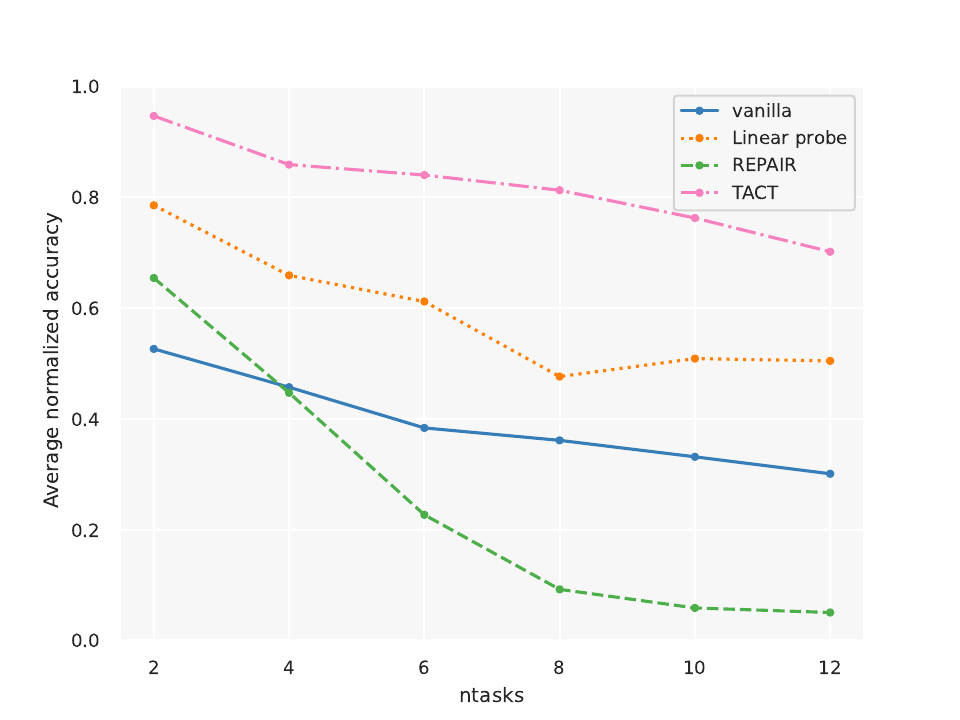}
    \caption{\Nonlocal merging ablation: We plot the average normalized accuracy across tasks (y-axis) versus the number of tasks merged (x-axis) different merging methods (VGG16).\textbf{(Left)} TACT: Applying TACT significantly improves the performance when compared to the baseline \nonlocal merging;  \textbf{(Center)} REPAIR: Applying REPAIR improves the  baseline \nonlocal merging when merging few tasks but the performance deteriorates when we merge more number of models   ; \textbf{(Right)} Linear Probing: Linear probing the final layer leads to moderate gains to merging method but still under performs when compared to TACT.}
    \label{fig:tact-repair-lp-ablation}
\end{figure}

In \Cref{tab:non-local-merging}, we report the average accuracy of the merged model normalized by the expert's accuracy for VGG16. 
First we note that applying TACT on any merging method significantly improves its performance, even when the method is combined with TALL mask. We also note that nor REPAIR neither permutations alone do not bring the same performance improvements.
Finally, we note that TACT combined with TALL mask and task arithmetic outperforms all other merging methods. These results are consistent across all settings.

\paragraph{REPAIR for \nonlocal model merging.} In \Cref{fig:tact-repair-lp-ablation}(center) we compare the performance of applying REPAIR to the merged model obtained from vanilla \nonlocal model merging, where the repaired BatchNorm parameters are obtained by computing the activation statistics of each of the expert model on the mixture dataset, obtained by sampling equal amounts of data from each of the task specific dataset, $\calD_t$. We observe that REPAIR outperforms vanilla model merging when merging 4 tasks. However, as the number of tasks increase, the performance of REPAIR deteriorates drastically. Comparing this to \Cref{fig:tact-repair-lp-ablation} (left) where we apply TACT, we do not see such degradation.

\paragraph{Linear probing the merged model.} 
\label{sec:lp-analysis}
In \Cref{fig:tact-repair-lp-ablation} (right), we test if the poor performance of vanilla \nonlocal model merging is caused by incompatibility of penultimate layer representations with the expert classifier (which is fixed). We test this by training a new linear probe for the merged model.
We find that linear probing does indeed increase the performance of the merged model but the gains remain small when compared to TACT.

\subsection{Local model merging with TACT}
\label{sec:local-merging}

\begin{figure}[t]

    \centering 
    \includegraphics[width=0.32\linewidth]{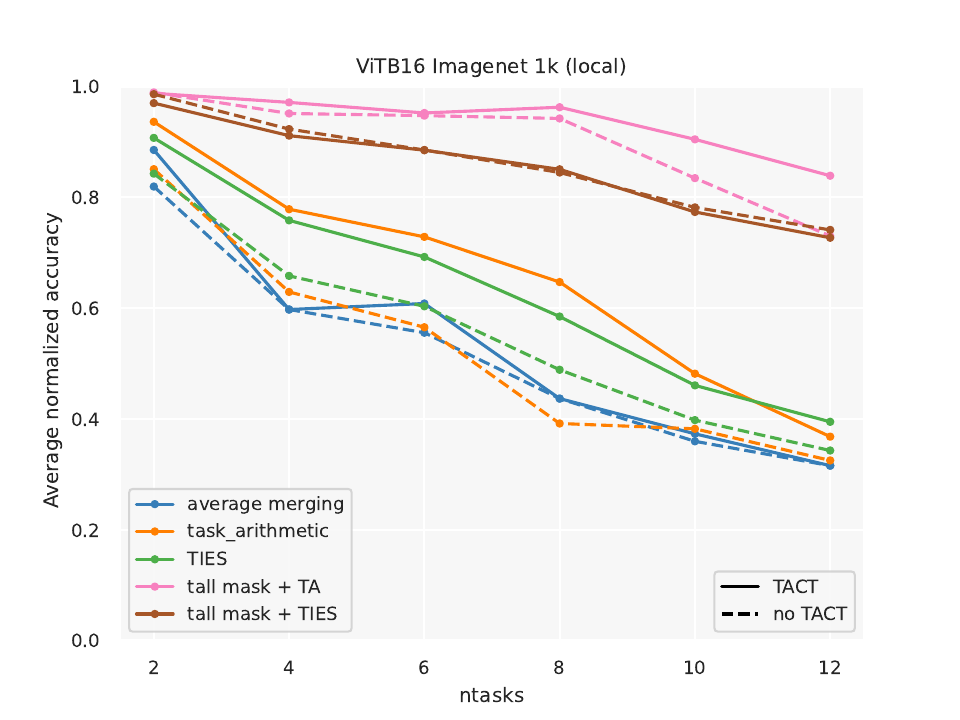}
    \includegraphics[width=0.32\linewidth]{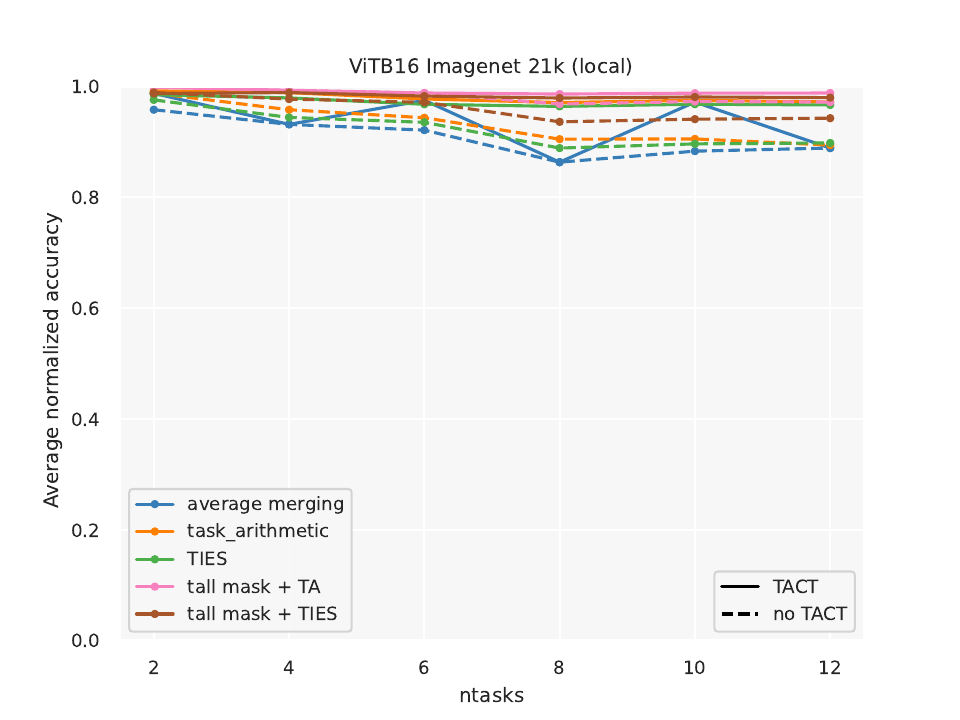}
    \includegraphics[width=0.32\linewidth]{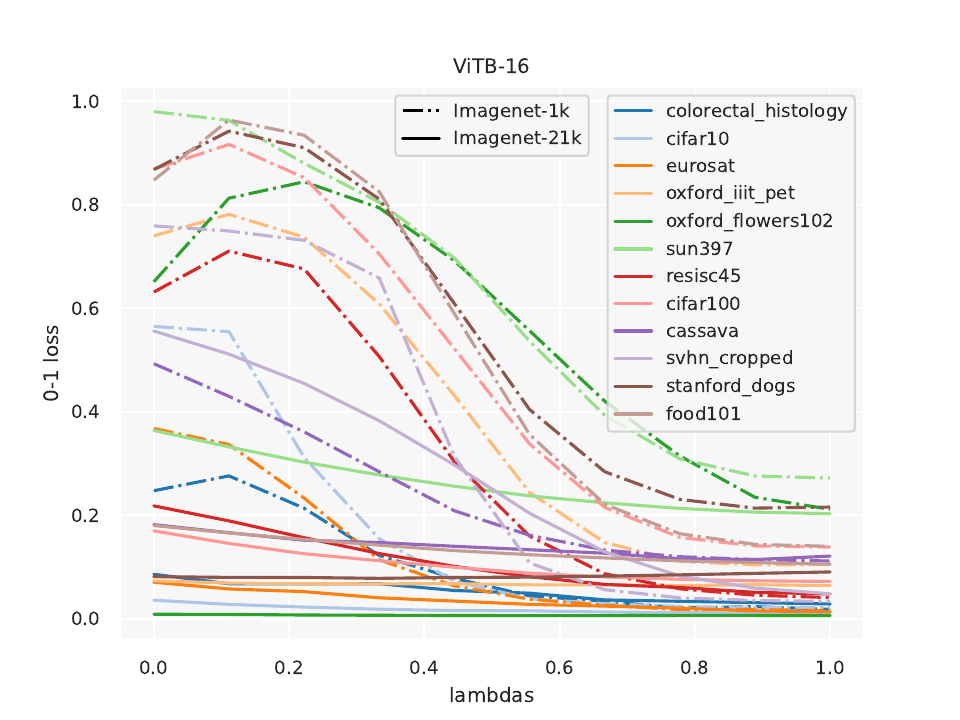}
    \caption{Local merging ViT-B16: First, we plot the average normalized accuracy across tasks (y-axis) and the number of tasks merged (x-axis): \textbf{(Left)} Local merging on ViTB16 pretrained with Imagenet-1k; \textbf{(Center)} Local merging on  ViTB16 pretrained with Imagenet-21k. \textbf{(Right)} Second, we plot the task specific 0--1 loss (y-axis) when interpolating between the initialization and the expert model (x-axis);}
\label{fig:local-avg-accuracy-v-ntasks}
\end{figure}

Next, we demonstrate that TACT improves merged model performance not only in the \nonlocal setting, but also in a local setting, where all experts were fine-tuned from the same foundational model.
We also evaluate the role of the foundation model when doing model merging. 

\paragraph{Setup.} We evaluate local merging in VGG16 model architecture pretrained on ImageNet-1k dataset, ViTB-16 model architecture pretrained on Imagnet-1k dataset, and ViTB-16 model architecture pretrained on ImageNet-21K dataset.
Note that ImageNet-21k is substantially bigger that Imagenet-1k with ~14 million images as compared to 1.2 million images. For each of the three foundation models, we derive experts on 12 tasks mentioned earlier.

\paragraph{Results.}  The results in \Cref{tab:local-merging} show that TACT improves model merging for every single method across all settings. 
Similar to \nonlocal merging, we find that TACT combined with TALL mask and task arithmetic retains most of the performance of the task-specific expert model.

\paragraph{Role of foundation model.} 
We investigate how the quality of the foundation model influences model merging.
\Cref{fig:local-avg-accuracy-v-ntasks} (left and center) reveals that weaker foundation models, i.e., the ones that were trained on smaller datasets, present significant deterioration in model merging results for standard baselines: Task Arithmetic and TIES merging (cf. \Cref{tab:local-merging} VGG16, ViTB16 (IN-1k)). 
In \Cref{fig:local-avg-accuracy-v-ntasks} (right), we observe that the expert models obtained from a weak foundation model, i.e. pretrained on the smaller Imagenet-1k dataset have an error barrier from initialization. This shows that experts obtained from weak foundation models tend to escape the local neighborhood of the pretrained model, thus increasing the interference between different task vectors. We see that applying TALL-mask merging together with TACT resolves this interference and outperforms other benchmarks. 
\begin{table}[]

\centering
\caption{Quantitative results for local model merging. We report average normalized accuracy for VGG16 model pretrained with Imagenet-1k, ViT-B16 model pretrained with Imagenet 1k, and ViTB-16 model pretrained with Imagenet-21k. }
\label{tab:local-merging}
\resizebox{0.9\linewidth}{!}{
\begin{tabular}{@{}cccccccccccccccccccccccccc@{}}
\toprule
\multicolumn{2}{c}{} &  & \multicolumn{3}{c}{VGG-16 (IN-1k)} & &  \multicolumn{1}{c}{} & & \multicolumn{3}{c}{ViT-B/16 (IN-1k)} & &  \multicolumn{1}{c}{} & & \multicolumn{3}{c}{ViT-B/16 (IN-21K)}\\ 
\cmidrule(l){3-7} \cmidrule(l){9-13} \cmidrule(l){15-19}
\multicolumn{1}{c}{} &  & \multicolumn{1}{c}{4 tasks} &  & \multicolumn{1}{c}{8 tasks} &  & \multicolumn{1}{c}{12 tasks} & &  \multicolumn{1}{c}{4 tasks} &  & \multicolumn{1}{c}{8 tasks} &  & \multicolumn{1}{c}{12 tasks} & & \multicolumn{1}{c}{4 tasks} &  & \multicolumn{1}{c}{8 tasks} &  & \multicolumn{1}{c}{12 tasks} &\\ 
\midrule

&Weight averaging &  70.88\% &  & 59.00\% &   & 54.43\% &  &  59.31\% &  & 43.70\% &  & 31.61\% &  & 93.15\% &  & 86.32\% & & 88.88\% \\

&  Task arithmetic &  72.99\% &  & 59.72\% &  & 55.36\% &  &  62.93\% &  & 37.44\% &  & 32.53\% &  & 95.79\% &  & 90.48\% & & 89.41\%\\

&  TIES &  76.48\% &  & 65.87\% &   & 61.27\% &  &  64.99\% &  & 48.18\% &  & 34.32\% &  &94.40\%&  & 88.87\% & & 89.79\%\\

 & {TALL Mask + TA } &  95.77\% &  & 94.40\% &   & 89.81\% &  &  94.70\% &  & 88.36\% &  & 73.01\% &  & 99.22\% &  & 96.81\% &  & 97.17\% \\

\multirow{-6}{*}{\rotatebox[origin=c]{90}{Vanilla}} &
  {TALL Mask + TIES} &  96.26\% &  & 88.39\% &   & 82.49\% &  &  91.33\% &  & 82.94\% &  & 74.11\% &  &97.71\%&  & 93.60\% & & 93.29\%\\ 
 \midrule
 &Weight averaging &  88.94\% &  & 81.93\% &   & 73.41\% &  &  70.55\% &  & 48.80\% &  & 31.05\% &  & 98.31\% & & 96.62\% & & 97.11\%\\

&  Task arithmetic &  91.03\% &  &84.67\%  &   & 76.93\% &  &  79.40\% &  & 62.10\% &  & 32.53\% &  & 98.88\% &  & 97.08\% &  & 97.19\%\\

&  TIES &  91.37\% &  & 84.96\% &   & 76.29\% &  &  74.75\% &  & 57.35\% &  & 34.32\% &  & 97.92\% &  &96.39\%  & & 96.65\%\\

 &  {TALL Mask + TA} &  \textbf{97.40\%} &  & \textbf{95.94\%} &   & \textbf{92.09\%} &  &  \textbf{97.11\%} &  & \textbf{94.03\%} &  & \textbf{83.88\%} &  & \textbf{99.30\%} &  & \textbf{98.62\%} & & \textbf{98.80\%}\\

\multirow{-6}{*}{\rotatebox[origin=c]{90}{TACT}} &
  {TALL Mask + TIES} &  96.37\% &  & 90.73\% &   & 84.55\% &  & 90.61\% &  & 83.55\% &  & 72.70\% &  & 98.89\% &  & 97.90\% &  & 97.98\%\\ 
 
 \bottomrule
\end{tabular}
}
\end{table}

\section{Conclusion}

We investigated the challenge of \nonlocal model merging, where expert models are not derived from the same foundation model. 
We demonstrated that while permutation alignment enables the merging of such models, it can lead to significant discrepancies in activation statistics, hindering overall performance. 

To address this, we introduced Task-specific Activation Correction (TACT), a technique that employs layer-wise rescaling and shifting to align the activations of the merged model with those of the expert models. 
Our experiments across various architectures and tasks showcased the effectiveness of TACT in significantly improving the performance of non-local model merging, with minimal increase in memory footprint.
By enabling the efficient merging of expert models from diverse origins, this research opens up new possibilities for model reuse and deployment.

While our proposed TACT method demonstrates promising results in \nonlocal model merging, there are limitations and avenues for future research.
Firstly, TACT relies on task-specific correction, requiring the computation of activation statistics for each task. Exploring data-agnostic correction methods that generalize across tasks could further simplify the merging process and reduce computational overhead. Secondly, our experiments primarily focus on expert models derived from foundation models amenable to existing permutation algorithms. %

Finally, merging experts derived from multiple different foundation models would require the so-called \emph{strong linear connectivity} to hold for SGD-trained networks. 
Strong linear connectivity modulo permutation for SGD-trained networks, as defined by \citet{sharma2024simultaneous}, would mean that multiple models can be permuted once in a way that they all end up being linearly connected. However, as highlighted in the same paper, strong linear connectivity has not been theoretically or empirically established  in the literature. 
As a result, our experiments are restricted to expert models originating from \emph{only two} foundation models. 
Future work could explore merging models from a more diverse set of foundation models as advances in permutation algorithms and theoretical understanding of connectivity emerge.

\section*{Reproducibility}
We release the code and model checkpoints used in our paper. Our code and checkpoints can be found at \url{https://github.com/ekanshs/tact-merge}. We also report all the hyperparameters used for training and merging in \cref{tab:ft-params} and \cref{tab:merging-hyperparams} respectively.

\section*{Acknowledgments}
ES is supported by the Vector Research Grant at the Vector Institute. DMR is supported by the funding through NSERC Discovery Grant and 
Canada CIFAR AI Chairs at the Vector Institute. The authors are grateful for the technical support from the Vector teams in maintaining the Vector Compute Cluster. Resources used to prepare this research were provided in part by the Vector Institute (vectorinstitute.ai), the Province of Ontario, the Government of Canada through CIFAR, and companies sponsoring the Vector Institute (\url{https://vectorinstitute.ai/partnerships/current-partners/})

\bibliography{iclr2025_conference}

\begin{thebibliography}{30}
\providecommand{\natexlab}[1]{#1}
\providecommand{\url}[1]{\texttt{#1}}
\expandafter\ifx\csname urlstyle\endcsname\relax
  \providecommand{\doi}[1]{doi: #1}\else
  \providecommand{\doi}{doi: \begingroup \urlstyle{rm}\Url}\fi

\bibitem[Ainsworth et~al.(2023)Ainsworth, Hayase, and Srinivasa]{ainsworth2022git}
Samuel Ainsworth, Jonathan Hayase, and Siddhartha Srinivasa.
\newblock Git re-basin: Merging models modulo permutation symmetries.
\newblock In \emph{The Eleventh International Conference on Learning Representations}, 2023.
\newblock URL \url{https://openreview.net/forum?id=CQsmMYmlP5T}.

\bibitem[Bossard et~al.(2014)Bossard, Guillaumin, and Van~Gool]{bossard14}
Lukas Bossard, Matthieu Guillaumin, and Luc Van~Gool.
\newblock Food-101 -- mining discriminative components with random forests.
\newblock In David Fleet, Tomas Pajdla, Bernt Schiele, and Tinne Tuytelaars (eds.), \emph{Computer Vision -- ECCV 2014}, pp.\  446--461, Cham, 2014. Springer International Publishing.

\bibitem[Cheng et~al.(2017)Cheng, Han, and Lu]{cheng17resisc}
Gong Cheng, Junwei Han, and Xiaoqiang Lu.
\newblock Remote sensing image scene classification: Benchmark and state of the art.
\newblock \emph{Proceedings of the IEEE}, 105\penalty0 (10):\penalty0 1865--1883, 2017.
\newblock \doi{10.1109/JPROC.2017.2675998}.

\bibitem[Dosovitskiy et~al.(2021)Dosovitskiy, Beyer, Kolesnikov, Weissenborn, Zhai, Unterthiner, Dehghani, Minderer, Heigold, Gelly, Uszkoreit, and Houlsby]{dosovitskiy2021an}
Alexey Dosovitskiy, Lucas Beyer, Alexander Kolesnikov, Dirk Weissenborn, Xiaohua Zhai, Thomas Unterthiner, Mostafa Dehghani, Matthias Minderer, Georg Heigold, Sylvain Gelly, Jakob Uszkoreit, and Neil Houlsby.
\newblock An image is worth 16x16 words: Transformers for image recognition at scale.
\newblock In \emph{International Conference on Learning Representations}, 2021.
\newblock URL \url{https://openreview.net/forum?id=YicbFdNTTy}.

\bibitem[Entezari et~al.(2022)Entezari, Sedghi, Saukh, and Neyshabur]{entezari2021role}
Rahim Entezari, Hanie Sedghi, Olga Saukh, and Behnam Neyshabur.
\newblock The role of permutation invariance in linear mode connectivity of neural networks.
\newblock In \emph{International Conference on Learning Representations}, 2022.
\newblock URL \url{https://openreview.net/forum?id=dNigytemkL}.

\bibitem[ErnestMwebaze(2019)]{cassava-disease}
Timnit~Gebru ErnestMwebaze.
\newblock Cassava disease classification, 2019.
\newblock URL \url{https://kaggle.com/competitions/cassava-disease}.

\bibitem[Frankle \& Carbin(2019)Frankle and Carbin]{frankle2018lottery}
Jonathan Frankle and Michael Carbin.
\newblock The lottery ticket hypothesis: Finding sparse, trainable neural networks.
\newblock In \emph{International Conference on Learning Representations}, 2019.
\newblock URL \url{https://openreview.net/forum?id=rJl-b3RcF7}.

\bibitem[Frankle et~al.(2020)Frankle, Dziugaite, Roy, and Carbin]{frankle2020linear}
Jonathan Frankle, Gintare~Karolina Dziugaite, Daniel Roy, and Michael Carbin.
\newblock Linear mode connectivity and the lottery ticket hypothesis.
\newblock In \emph{Proceedings of the 37th International Conference on Machine Learning}, volume 119, pp.\  3259--3269. PMLR, 2020.
\newblock URL \url{https://proceedings.mlr.press/v119/frankle20a.html}.

\bibitem[He et~al.(2022)He, Chen, Xie, Li, Dollár, and Girshick]{kaiming22vitmae}
Kaiming He, Xinlei Chen, Saining Xie, Yanghao Li, Piotr Dollár, and Ross Girshick.
\newblock Masked autoencoders are scalable vision learners.
\newblock In \emph{2022 IEEE/CVF Conference on Computer Vision and Pattern Recognition (CVPR)}, pp.\  15979--15988, 2022.
\newblock \doi{10.1109/CVPR52688.2022.01553}.

\bibitem[Helber et~al.(2019)Helber, Bischke, Dengel, and Borth]{helber_eurosat_2019}
Patrick Helber, Benjamin Bischke, Andreas Dengel, and Damian Borth.
\newblock Eurosat: A novel dataset and deep learning benchmark for land use and land cover classification.
\newblock \emph{IEEE Journal of Selected Topics in Applied Earth Observations and Remote Sensing}, 12\penalty0 (7):\penalty0 2217--2226, 2019.
\newblock \doi{10.1109/JSTARS.2019.2918242}.

\bibitem[Ilharco et~al.(2023)Ilharco, Ribeiro, Wortsman, Schmidt, Hajishirzi, and Farhadi]{ilharco2023editing}
Gabriel Ilharco, Marco~Tulio Ribeiro, Mitchell Wortsman, Ludwig Schmidt, Hannaneh Hajishirzi, and Ali Farhadi.
\newblock Editing models with task arithmetic.
\newblock In \emph{The Eleventh International Conference on Learning Representations}, 2023.
\newblock URL \url{https://openreview.net/forum?id=6t0Kwf8-jrj}.

\bibitem[Jordan et~al.(2023)Jordan, Sedghi, Saukh, Entezari, and Neyshabur]{jordan2022repair}
Keller Jordan, Hanie Sedghi, Olga Saukh, Rahim Entezari, and Behnam Neyshabur.
\newblock {REPAIR}: {RE}normalizing {P}ermuted {A}ctivations for {I}nterpolation {R}epair.
\newblock In \emph{The Eleventh International Conference on Learning Representations}, 2023.
\newblock URL \url{https://openreview.net/forum?id=gU5sJ6ZggcX}.

\bibitem[Kather et~al.(2016)Kather, Zöllner, Bianconi, Melchers, Schad, Gaiser, Marx, and Weis]{kather_2016_53169}
Jakob~Nikolas Kather, Frank~Gerrit Zöllner, Francesco Bianconi, Susanne~M Melchers, Lothar~R Schad, Timo Gaiser, Alexander Marx, and Cleo-Aron Weis.
\newblock {Collection of textures in colorectal cancer histology}, June 2016.
\newblock URL \url{https://doi.org/10.5281/zenodo.53169}.

\bibitem[Khosla et~al.(2011)Khosla, Jayadevaprakash, Yao, and Fei-Fei]{KhoslaYaoJayadevaprakashFeiFei_FGVC2011}
Aditya Khosla, Nityananda Jayadevaprakash, Bangpeng Yao, and Li~Fei-Fei.
\newblock Novel dataset for fine-grained image categorization.
\newblock In \emph{First Workshop on Fine-Grained Visual Categorization, IEEE Conference on Computer Vision and Pattern Recognition}, Colorado Springs, CO, June 2011.

\bibitem[Krizhevsky(2009)]{krizhevsky2009learning}
Alex Krizhevsky.
\newblock Learning multiple layers of features from tiny images.
\newblock \emph{Technical Report}, 2009.
\newblock URL \url{https://www.cs.toronto.edu/~kriz/learning-features-2009-TR.pdf}.

\bibitem[Nagarajan \& Kolter(2019)Nagarajan and Kolter]{nagarajan2019uniform}
Vaishnavh Nagarajan and J.~Zico Kolter.
\newblock Uniform convergence may be unable to explain generalization in deep learning.
\newblock In \emph{Advances in Neural Information Processing Systems}, volume~32, pp.\  11615--11626, 2019.
\newblock URL \url{https://proceedings.neurips.cc/paper/2019/hash/05e97c207235d63ceb1db43c60db7bbb-Abstract.html}.

\bibitem[Netzer et~al.(2011)Netzer, Wang, Coates, Bissacco, Wu, and Ng]{netzer11}
Yuval Netzer, Tao Wang, Adam Coates, Alessandro Bissacco, Bo~Wu, and Andrew Ng.
\newblock Reading digits in natural images with unsupervised feature learning.
\newblock \emph{NIPS}, 01 2011.

\bibitem[Parkhi et~al.(2012)Parkhi, Vedaldi, Zisserman, and Jawahar]{parkhi12a}
Omkar~M. Parkhi, Andrea Vedaldi, Andrew Zisserman, and C.~V. Jawahar.
\newblock Cats and dogs.
\newblock In \emph{IEEE Conference on Computer Vision and Pattern Recognition}, 2012.

\bibitem[Ridnik et~al.(2021)Ridnik, Ben-Baruch, Noy, and Zelnik-Manor]{ridnik2021imagenet21k}
Tal Ridnik, Emanuel Ben-Baruch, Asaf Noy, and Lihi Zelnik-Manor.
\newblock Imagenet-21k pretraining for the masses, 2021.

\bibitem[Russakovsky et~al.(2015)Russakovsky, Deng, Su, Krause, Satheesh, Ma, Huang, Karpathy, Khosla, Bernstein, Berg, and Fei-Fei]{ILSVRC15}
Olga Russakovsky, Jia Deng, Hao Su, Jonathan Krause, Sanjeev Satheesh, Sean Ma, Zhiheng Huang, Andrej Karpathy, Aditya Khosla, Michael Bernstein, Alexander~C. Berg, and Li~Fei-Fei.
\newblock {ImageNet Large Scale Visual Recognition Challenge}.
\newblock \emph{International Journal of Computer Vision (IJCV)}, 115\penalty0 (3):\penalty0 211--252, 2015.
\newblock \doi{10.1007/s11263-015-0816-y}.

\bibitem[Sharma et~al.(2024)Sharma, Kwok, Denton, Roy, Rolnick, and Dziugaite]{sharma2024simultaneous}
Ekansh Sharma, Devin Kwok, Tom Denton, Daniel~M. Roy, David Rolnick, and Gintare~Karolina Dziugaite.
\newblock Simultaneous linear connectivity of neural networks modulo permutation.
\newblock In \emph{Machine Learning and Knowledge Discovery in Databases. Research Track}, pp.\  262--279, Cham, 2024. Springer Nature Switzerland.
\newblock URL \url{https://doi.org/10.1007/978-3-031-70368-3_16}.

\bibitem[Simonyan \& Zisserman(2015)Simonyan and Zisserman]{Simonyan15}
Karen Simonyan and Andrew Zisserman.
\newblock Very deep convolutional networks for large-scale image recognition.
\newblock In \emph{International Conference on Learning Representations}, 2015.

\bibitem[Singh \& Jaggi(2020)Singh and Jaggi]{singh2020model}
Sidak~Pal Singh and Martin Jaggi.
\newblock Model fusion via optimal transport.
\newblock In \emph{Advances in Neural Information Processing Systems}, volume~33, pp.\  22045--22055, 2020.
\newblock URL \url{https://proceedings.neurips.cc/paper/2020/hash/fb2697869f56484404c8ceee2985b01d-Abstract.html}.

\bibitem[Wang et~al.(2020)Wang, Yurochkin, Sun, Papailiopoulos, and Khazaeni]{wang2020federated}
Hongyi Wang, Mikhail Yurochkin, Yuekai Sun, Dimitris Papailiopoulos, and Yasaman Khazaeni.
\newblock Federated learning with matched averaging.
\newblock \emph{arXiv preprint arXiv:2002.06440}, 2020.

\bibitem[Wang et~al.(2024)Wang, Dimitriadis, Ortiz-Jimenez, Fleuret, and Frossard]{wang2024localizing}
Ke~Wang, Nikolaos Dimitriadis, Guillermo Ortiz-Jimenez, Fran{\c{c}}ois Fleuret, and Pascal Frossard.
\newblock Localizing task information for improved model merging and compression.
\newblock In \emph{Forty-first International Conference on Machine Learning}, 2024.
\newblock URL \url{https://openreview.net/forum?id=DWT9uiGjxT}.

\bibitem[Wolf et~al.(2020)Wolf, Debut, Sanh, Chaumond, Delangue, Moi, Cistac, Rault, Louf, Funtowicz, Davison, Shleifer, von Platen, Ma, Jernite, Plu, Xu, Le~Scao, Gugger, Drame, Lhoest, and Rush]{wolf-etal-2020-transformers}
Thomas Wolf, Lysandre Debut, Victor Sanh, Julien Chaumond, Clement Delangue, Anthony Moi, Pierric Cistac, Tim Rault, Remi Louf, Morgan Funtowicz, Joe Davison, Sam Shleifer, Patrick von Platen, Clara Ma, Yacine Jernite, Julien Plu, Canwen Xu, Teven Le~Scao, Sylvain Gugger, Mariama Drame, Quentin Lhoest, and Alexander Rush.
\newblock Transformers: State-of-the-art natural language processing.
\newblock In Qun Liu and David Schlangen (eds.), \emph{Proceedings of the 2020 Conference on Empirical Methods in Natural Language Processing: System Demonstrations}, pp.\  38--45, Online, October 2020. Association for Computational Linguistics.
\newblock \doi{10.18653/v1/2020.emnlp-demos.6}.
\newblock URL \url{https://aclanthology.org/2020.emnlp-demos.6}.

\bibitem[Xiao et~al.(2010)Xiao, Hays, Ehinger, Oliva, and Torralba]{xiao2010sun}
Jianxiong Xiao, James Hays, Krista~A. Ehinger, Aude Oliva, and Antonio Torralba.
\newblock Sun database: Large-scale scene recognition from abbey to zoo.
\newblock In \emph{2010 IEEE Computer Society Conference on Computer Vision and Pattern Recognition}, pp.\  3485--3492, 2010.
\newblock \doi{10.1109/CVPR.2010.5539970}.

\bibitem[Yadav et~al.(2023)Yadav, Tam, Choshen, Raffel, and Bansal]{yadav2023tiesmerging}
Prateek Yadav, Derek Tam, Leshem Choshen, Colin Raffel, and Mohit Bansal.
\newblock {TIES}-merging: Resolving interference when merging models.
\newblock In \emph{Thirty-seventh Conference on Neural Information Processing Systems}, 2023.
\newblock URL \url{https://openreview.net/forum?id=xtaX3WyCj1}.

\bibitem[Yadav et~al.(2024)Yadav, Tam, Choshen, Raffel, and Bansal]{yadav2024ties}
Prateek Yadav, Derek Tam, Leshem Choshen, Colin~A Raffel, and Mohit Bansal.
\newblock Ties-merging: Resolving interference when merging models.
\newblock \emph{Advances in Neural Information Processing Systems}, 36, 2024.

\bibitem[Zhu \& Gupta(2017)Zhu and Gupta]{zhu2017prune}
Michael Zhu and Suyog Gupta.
\newblock To prune, or not to prune: exploring the efficacy of pruning for model compression.
\newblock \emph{arXiv preprint arXiv:1710.01878}, 2017.

\end{thebibliography}
\bibliographystyle{iclr2025_conference}

\clearpage

\appendix

\section{Experimental details}
\label{app:exp-details}
For our experiments we use convolutional neural network, VGG16, with LayerNormalization layer after each convolutional layer each, and a standard transformer based neural vision network, ViT-B16. 

We pretrain the VGG16 model architecture using SGD with momentum with learning rate of 0.1 and momentum of 0.9, using a batch size of 512. For ViT-B16 foundation models, we use the checkpoint released by \cite{wolf-etal-2020-transformers} which is pre-trained on Imagenet-21k, and the checkpoint released by \citet{kaiming22vitmae} which is pretrained on Imagenet-1k. 

For our evaluation we finetune the pretrained model the following classification datasets: CIFAR-10, CIFAR-100, SUN397, Cassava, Flowers102, EuroSAT, RESISC45, Colorectal Histology, Stanford Dogs, Oxford IIIT Pet, Food101, and SVHN. For each foundation model we report the finetuning hyperparameters in \Cref{tab:ft-params}. We report the accuracies obtained by the expert models in \Cref{tab:expert-accs}

\subsection{\Nonlocal model merging}

For evaluation in the \nonlocal setting, we split the tasks in two buckets that: 
\begin{enumerate}
    \item Tasks A: Colorectal Histology, CIFAR10, Eurosat,  Oxford IIIT Pet, Oxford Flowers, SUN397
     \item Tasks B: RESISC, CIFAR100, Cassava, and  SVHN, Stanford Dogs, Food101
\end{enumerate}

To evaluate in the \nonlocal setting we pick equal number of tasks from each bucket and run the non-local merging algorithm \cref{alg:nlocal-merging}. For example, for merging 4 models, we select two tasks from Tasks A bucket and two tasks from Tasks B bucket and evaluate the performance of model merging. 
We search through following hyperparameters on a small validation set obtained from the test set for model merging:
\begin{table}[!]
    \centering
\resizebox{0.9\linewidth}{!}{
    \begin{tabular}{c|cccccccccc}
    \toprule
     & 2 tasks & 4 tasks & 6 tasks  & 8 tasks  & 10 tasks & 12 tasks \\
    \midrule
        Task Arithmetic  &  $\lambda\in\set{0.5, 0.7, 1.0}$ & $\lambda\in\set{0.4, 0.6, 0.9}$ &  $\lambda\in\set{0.3, 0.5, 0.8}$ & $\lambda\in\set{0.2, 0.5 ,0.7}$ & $\lambda\in\set{0.2, 0.4, 0.6}$ & $\lambda\in\set{0.1,0.3, 0.5}$\\
        TIES  &  $\lambda\in\set{0.3, 0.5, 0.7, 1.0}$ & $\lambda\in\set{0.3, 0.5, 0.7, 1.0}$ &  $\lambda\in\set{0.3, 0.5, 0.7, 1.0}$ & $\lambda\in\set{0.3, 0.5, 0.7, 1.0}$ & $\lambda\in\set{0.3, 0.5, 0.7, 1.0}$ & $\lambda\in\set{0.3, 0.5, 0.7, 1.0}$\\
        TA + TALL  &  $\lambda\in\set{0.5, 0.7, 1.0}$ & $\lambda\in\set{0.4, 0.6, 0.9}$ &  $\lambda\in\set{0.3, 0.5, 0.8}$ & $\lambda\in\set{0.2, 0.5 ,0.7}$ & $\lambda\in\set{0.2, 0.4, 0.6}$ & $\lambda\in\set{0.1,0.3, 0.5}$\\
        TIES + TALL rate &  $\lambda\in\set{0.3, 0.5, 0.7, 1.0}$ & $\lambda\in\set{0.3, 0.5, 0.7, 1.0}$ &  $\lambda\in\set{0.3, 0.5, 0.7, 1.0}$ & $\lambda\in\set{0.3, 0.5, 0.7, 1.0}$ & $\lambda\in\set{0.3, 0.5, 0.7, 1.0}$ & $\lambda\in\set{0.3, 0.5, 0.7, 1.0}$\\
    \bottomrule
    \end{tabular}
    }
    \caption{Merging hyperparameters}
    \label{tab:merging-hyperparams}
\end{table}

These evaluations are reported in \Cref{tab:non-local-merging}.

\begin{table}[]
    \centering

    \begin{tabular}{c|cccc}
    \toprule
     & VGG16 & ViT-B16 (IN-1k) & ViT-B16 (IN-21k) \\
    \midrule
        Steps  &  8000 & 10000 &  10000\\
        Linear warmup steps  & 500  & 500 & 500 \\
        Learning schedule  & cosine decay  & cosine decay & cosine decay \\
        Peak learning rate &  0.01 & 0.01 & 0.001 \\
        Weight decay &  0.001 & 0.1  & 0.1 \\
    \bottomrule
    \end{tabular}
    
    \caption{Finetuning hyperparameters}
    \label{tab:ft-params}
\end{table}

\begin{table}[]
    \centering
\resizebox{0.99\linewidth}{!}{
    \begin{tabular}{ccccccccccccccccccccc}
    \toprule
    Models & Cassava &  CIFAR10 & CIFAR100 & Colorectal Histology & EuroSat & Food101 & Oxford Flowers & Oxford IIIT Pet & RESISC & Stanford Dogs & SUN397 & SVHN \\
    \midrule
        VGG16  & 85.04\%  & 96.90\% & 82.35\% & 96.88\% & 98.47\% & 76.81\% & 78.43\% & 86.53\% & 93.16\% & 74.68\% & 66.18\% &95.09\% \\
        ViT-B16 (IN-1k)  & 88.84\%  & 97.86\% & 86.13\% & 98.17\% & 98.43\% & 86.00\% & 78.84\% & 89.53\% &95.90\% & 78.41\% & 72.81\% &96.64\% \\
        ViT-B16 (IN-21k) &  87.89\% & 98.86\% & 92.77\% & 97.14\% & 98.66\% & 89.45\% & 99.38\% & 93.59\% &95.18\% & 90.95\% & 79.70\%  & 95.17\% \\
        
    \bottomrule
    \end{tabular}
    }
    \caption{Expert accuracies}
    \label{tab:expert-accs}
\end{table}

\section{Aligning networks with permutation}
\label{app:weightmatching}
We use the weight matching algorithm to align the weights of the two pretrained foundation models. 

For two networks, A and B, with identical architectures, weight matching algorithm operates by finding a permutation that $P$ that aims at minimizing the following objective:
\[
P^\star = \argmin_{P\in \calS} \norm{A - \permuteop{P}{B}}_2
\]

For deep neural networks, finding a permutation $P$ that exactly minimizes the objective is \np hard. \citet{ainsworth2022git}  instead propose a layer wise approximation to this minimization problem that works well in practice for wide neural networks. Their algorithm operates as follows:

Let $A^{(k)}, B^{(k)} \in \Reals^{p\times q}$ denote the weight matrix associated with the $k^{\text{th}}$ layer of the neural networks $A$ and $B$ respectively. Let $P^{(k)} \in \calS_p$ denote the permutation of the $k^\text{th}$ parameter set. We can optimize the permutation $P^{(k)}$ that minimizes the $\ell_2$ distance between $A^{(k)}, B^{(k)}$:
\begin{align*}
{P^{(k)}} &= \argmin_{P\in \calS_p} \norm{A^{(k)} - \permuteop{P}{B^{(k)}}}_2\\
    &= \argmax_{P\in \calS_p} \mrm{Trace} \left(\binner{A^{(k)}}{B^{(k)}} P^\top\right)
\end{align*}

The weight matching algorithm operates as follows:
\begin{enumerate}
    \item
    Randomly shuffle the order in which sets of permutations are solved.
    \item
    For each parameter set $k$,
        \begin{enumerate}
        \item 
        Compute $G = A^{(k)} B^{(k)\top}$.
        \item 
        Use a linear sum assignment algorithm to find a permutation which maximizes pairwise similarity:
        \[
        P^\star = \argmax_{P \in S_p} \{\mrm{Trace} (GP^T)\}.
        \]
        \item
        Apply the new permutation $P^\star$ to $B^{(k)}$ and $P^{(k)}$.
    \end{enumerate}
    \item
    Repeat until either pairwise similarity does not improve in every permutation set.
\end{enumerate}

\section{Task specific activation correction (TACT) algorithm}
\label{app:tact}
In this section we provide the details related to the TACT procedure we use to improve the performance of model merging.  Let $\calA^{(k)}_{\theta}(X)$ denote the channel activations of $k^{\text{th}}$ module for some neural network parameterized by $\theta$ on some input data $X$, and let $f^{(k)}_{\theta}(x)$ denote the computation done by the $k^{\text{th}}$ module. 

TACT correction for task $t$ is summarized as follows:
\begin{enumerate}
    \item Compute the activation statistics of the expert model:
    \[
    \mu^{(k)}_t = \EE{X\sim \calD_t}{\calA^{(k)}_{\theta_t}(X)}, \quad \sigma^{(k)}_t = \std_{X\sim \calD_t} [\calA^{(k)}_{\theta_t}(X)], \forall k\in [K].
    \]
    This can be done by adding BatchNorm layer to each module for which we need to compute these statistics and running a forward pass of the training data. 
    \item We update the merged model to include the BatchNorm after each module that needs correction. The parameters of the BatchNorm are set to be the statistics computed on the expert model. We summarize the forward pass through the merged model in \Cref{alg:merged-model-fp}. 
\begin{algorithm}[!h]
\caption{$\mrm{MergedModel}$}
\textbf{input:}  Data: $X$, Merged parameters: $\theta_\merged$, Activation Statistics: $(\mu_1, \sigma_1),\dots, (\mu_K, \sigma_k)$
    \begin{algorithmic}[1]
    \State $X_0 \gets X$
        \For{$k \in [1, \dots, K]$}  
        \State $X_k \gets f^{(k)}_{\theta}(X_{k-1})$
        \State $X_k \gets \mrm{BatchNorm}(\mu = \mu_k, \sigma = \sigma_k)$
        \EndFor
    \end{algorithmic}
\textbf{return:} $X_K$
\label{alg:merged-model-fp}
\end{algorithm}

    \item Finally, we do a forward pass through the training data $\traindata{t}$ to the merged model to update the internal statistics of the BatchNorm statistics. This updates model is used for inference on $\testdata{t}$
\end{enumerate}

\subsection{Cost}
\label{sec:cost}
We note that applying TACT comes with additional computation and memory requirements, which we analyze below.

\paragraph{Memory cost.} We examine the memory cost of deploying a merged model  with TACT. TACT needs to store additional task specific BatchNorm parameters and task specific statistics.  Thus the memory cost of deploying a merged model with TACT scales linearly with the number of tasks. However the memory footprint of storing BatchNorm parameters and statistics is very small when compared to storing the expert parameters. For instance, for ViT-B16 model architecture, the memory footprint of the storing a single expert is around 655 MB, however storing the BatchNorm parameters and statistics is less than 1MB.
Similarly for VGG16, the memory footprint of storing a single expert is around 625MB, whereas storing BatchNorm parameters and statistics is 0.3MB. 

\paragraph{Computational cost.} For TACT, we need a single forward pass through each expert to compute the statistics of the expert model. Also, we need a single forward pass of each of the training data through the merged model to update the task data-dependent batch statistics. Thus, for merging $T$ models we need $2 \cdot T$ forward passes to compute all the statistics needed for TACT correction.

\end{document}